\pdfoutput=1

\documentclass[11pt]{article}

\usepackage[final]{coling}
\usepackage{times}
\usepackage{bm}
\usepackage{latexsym}
\usepackage{amssymb}
\usepackage{booktabs}
\usepackage{multirow}
\usepackage{amsmath}
\usepackage{makecell}
\usepackage{amssymb}
\usepackage{subcaption}
\usepackage[T1]{fontenc}

\usepackage[utf8]{inputenc}

\usepackage{microtype}

\usepackage{inconsolata}

\usepackage{graphicx}

\usepackage{amsmath,amsfonts,bm}

\def\eqref#1{equation~\ref{#1}}

\def\1{\bm{1}}

\DeclareMathAlphabet{\mathsfit}{\encodingdefault}{\sfdefault}{m}{sl}
\SetMathAlphabet{\mathsfit}{bold}{\encodingdefault}{\sfdefault}{bx}{n}

\usepackage{amsmath}
\usepackage{amsthm}

\newtheorem{theorem}{Theorem}
\newtheorem{definition}{Definition}

\newtheorem{property}{Property}

\usepackage{enumitem}

\makeatletter
\newcommand{\rmnum}[1]{\romannumeral #1}
\newcommand{\Rmnum}[1]{\expandafter\@slowromancap\romannumeral #1@}
\makeatother

\title{Is Parameter Collision Hindering Continual Learning in LLMs?}

\author{
  Shuo Yang$^{1,*}$, Kun-Peng Ning$^{1,*}$, Yu-Yang Liu$^{1,\dagger}$, Jia-Yu Yao$^1$, \\
  \textbf{Yong-Hong Tian$^{1,2}$}, \textbf{Yibing Song$^{3,4}$}, \textbf{Li Yuan$^{1,2,\dagger}$} \\
  \textsuperscript{1}School of Electronic and Computer Engineering, Peking University \\
  \textsuperscript{2}Peng Cheng Laboratory\quad 
  \textsuperscript{3}DAMO Academy, Alibaba Group\quad
  \textsuperscript{4}Hupan Lab\\
  \texttt{\{shuo\_yang, ningkp\}@stu.pku.edu.cn}, \\
  \texttt{\{liuyuyang13, jiayu\_yao, yhtian, yuanli-ece\}@pku.edu.cn} \\
}

\begin{document}
\maketitle

\renewcommand{\thefootnote}{\relax}
\footnote{$^{*}$These authors contributed equally to this work.}
\footnote{$^{\dagger}$Corresponding authors.}
\footnote{The code and dataset are available at \url{https://github.com/PKU-YuanGroup/N-LoRA}.}
\renewcommand{\thefootnote}{\arabic{footnote}}


\begin{abstract}

Large Language Models (LLMs) often suffer from catastrophic forgetting when learning multiple tasks sequentially, making continual learning (CL) essential for their dynamic deployment. Existing state-of-the-art (SOTA) methods, such as O-LoRA, typically focus on constructing orthogonality tasks to decouple parameter interdependence from various domains.
In this paper, we reveal that building non-collision parameters is a more critical factor in addressing CL challenges. 
Our theoretical and experimental analyses demonstrate that non-collision parameters can provide better task orthogonality, which is a sufficient but unnecessary condition. Furthermore, knowledge from multiple domains will be preserved in non-collision parameter subspaces, making it more difficult to forget previously seen data. Leveraging this insight, we propose \textbf{N}on-collision \textbf{Lo}w-\textbf{R}ank \textbf{A}daptation (\textbf{N-LoRA}), a simple yet effective approach leveraging low collision rates to enhance CL in LLMs. Experimental results on multiple CL benchmarks indicate that N-LoRA achieves superior performance ($\bm{+2.9\%}$), higher task orthogonality ($\bm{\times4.1}$ \textbf{times}), and lower parameter collision ($\bm{\times58.1}$ \textbf{times}) than SOTA methods.



\end{abstract}

\section{Introduction}



Continual Learning (CL)~\cite{de2021continual,silver2013lifelong} is essential for adapting large language models (LLMs) to new tasks, but it often leads to \textit{catastrophic forgetting} \cite{mccloskey1989catastrophic}, where knowledge from previous tasks is overwritten. 
Various methods, such as rehearsal-based~\cite{rebuffi2017incremental}, regularization-based~\cite{kirkpatrick2017overcoming}, and architecture-based approaches~\cite{mallya2018packnet}, have been proposed to mitigate this, but they significantly increase computational demands due to the vast number of parameters in LLMs ~\cite{wu2024continual}.
Recent advances in parameter-efficient fine-tuning (PEFT), such as LoRA~\cite{hu2021lora}, introduce a minimal set of trainable parameters while maintaining the rest of the pre-trained LLM fixed.
However, the task-specific parameters may collide with each other, potentially overwriting earlier knowledge and thus leading to poor model performance.

\begin{figure*}[!ht]
  \includegraphics[width=2\columnwidth]{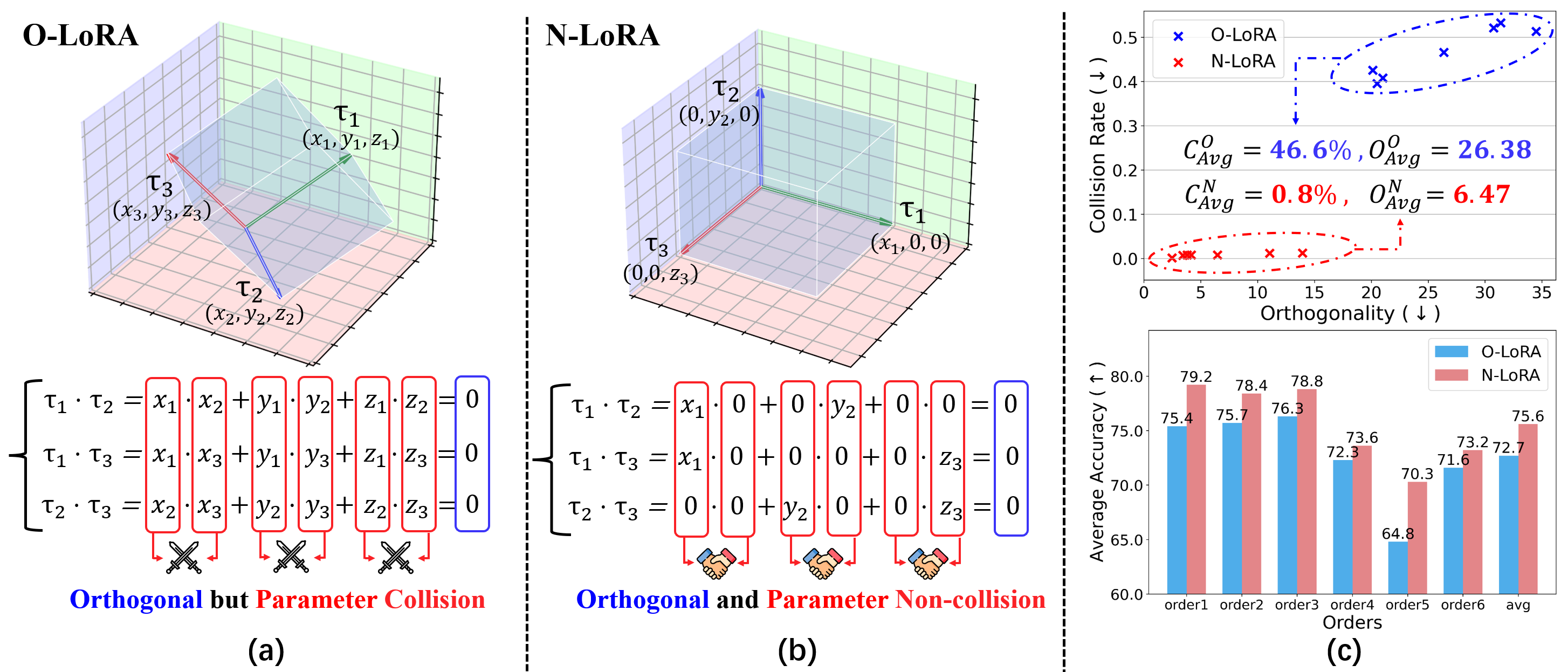}
  \caption{(a) Orthogonal but Parameter Collision: Tasks $\tau_1$, $\tau_2$, and $\tau_3$ are mutually orthogonal but interaction within each space, resulting in parameter collision. (b) Non-collision and orthogonal: Tasks $\tau_1$, $\tau_2$, and $\tau_3$ update only along distinct, non-conflicting subspaces, preserving prior task knowledge. (c) Performance Comparison: N-LoRA (red) and O-LoRA (blue) are compared across various metrics, with N-LoRA achieving lower collision rates, improved orthogonality, and superior average accuracy.}
  \label{fig:1}
\end{figure*}


To tackle this challenge in LLMs, existing studies~\cite{farajtabar2020orthogonal, saha2021gradient, wang2023orthogonal} attempted to employ orthogonal gradient descent-based methods to decouple parameter interdependence across different domains. Among them, O-LoRA (the current SOTA CL method in LLMs) incrementally updates task-specific parameters by constraining them to the orthogonal subspace of previous tasks, ensuring the independence of LoRA parameters across subspaces\cite{wang2023orthogonal}. However, despite this independence, interactions between LoRA parameters in different subspaces may still occur, potentially leading to parameter collisions.


In this paper, we reveal that constructing non-collision LoRA parameters provides a more effective solution to the problem of parameter interdependence, as non-collision serves as a sufficient but not necessary condition for orthogonality. 
By avoiding unnecessary overlaps in the parameter space, non-collision inherently minimizes interference among task-specific parameters, ensuring better task isolation and knowledge retention. 
As illustrated in Figure \ref{fig:1}(a), assuming a three-dimensional parameter space, existing methods (\textit{e.g.} O-LoRA) aim to learn three mutually orthogonal (blue box) task vectors $\tau_1$, $\tau_2$, and $\tau_3$. Each task vector $[x_i, y_i, z_i]$ occupies all three subspaces of the parameters, and these parameters will interact with each other (red box), hurting the model performance. 
In contrast, as illustrated in Figure \ref{fig:1}(b), non-collision constructs an extremely sparse parameter space, allowing each task vector to be updated in a local subspace without affecting each other. Non-collision parameters naturally achieve better task orthogonality (blue box), effectively decoupling parameter interdependence while preserving prior knowledge.

Leveraging the above insights, we propose \textbf{N}on-collision \textbf{Lo}w-\textbf{R}ank \textbf{A}daptation (\textbf{N-LoRA}), a simple yet effective approach to tackling continual learning problems in LLMs. Instead of directly optimizing the collision rate, we introduce $\ell_1$ constraints on task-specific LoRA parameters to make them extremely sparse, thereby significantly reducing the collision and dependency between various task vectors. 
This sparsity ensures that task-specific parameters operate independently in dedicated subspaces, avoiding interference and allowing for seamless task transitions. 
As illustrated in Figure \ref{fig:1}(c), compared to the orthogonal-based method (O-LoRA), the proposed non-collision-based approach (N-LoRA) can significantly achieve better task orthogonality ($\bm{\times4.1}$ \textbf{times}), lower parameter collision ($\bm{\times58.1}$ \textbf{times}), and superior average accuracy ($\bm{+2.9\%}$) in all cases.  



We empirically validate the effectiveness of the proposed N-LoRA through extensive experiments, demonstrating that it significantly outperforms prior SOTA methods in various datasets and model architectures. The contribution of this paper can be summarized as follows:

\begin{itemize}
    \item We reveal the intrinsic relationship between parameter orthogonality and collisions, clarifying that the key to addressing \textit{catastrophic forgetting} lies in preventing parameter collisions, not just enforcing orthogonality.
    \item We introduce a simple yet effective method, called N-LoRA, to tackle the CL problem in LLMs, which leverages low collision parameters to enhance the model performance.
    \item Our experiments on various CL benchmarks demonstrate that N-LoRA not only surpasses previous SOTA methods, but also achieves better orthogonality and a lower collision rate.
    \item N-LoRA is compatible with existing PEFT-based approaches and serves as a plug-and-play solution that can substantially enhance the performance of existing SOTA methods.
\end{itemize}

\section{Related Works}
\textbf{Parameter Efficient Fine-Tuning.}
Parameter Efficient Fine-Tuning (PEFT)~\cite{he2021towards, ding2022delta} has become a vital area of research aimed at enhancing the performance of large language models (LLMs) by adjusting only a small portion of the model's parameters. The primary objective of PEFT is to achieve performance on downstream tasks comparable to full model fine-tuning while minimizing computational resources by tuning only a small subset of model parameters. Various approaches have been proposed in this domain, including Adapter tuning~\cite{houlsby2019parameter}, Prompt-based tuning~\cite{wei2021finetuned}, P-tuning~\cite{liu2021p}, Prefix tuning~\cite{li2021prefix}, BitFit~\cite{zaken2021bitfit}, and LoRA~\cite{wang2023orthogonal}. In this work, we introduce an easy yet effective method built on LoRA to address this challenge, demonstrating strong performance.

\textbf{Conventional Continual Learning.} Conventional CL typically falls into three categories: (\rmnum{1}) Rehearsal-based approaches~\cite{li2017learning, liu2021adaptive, lopez2017gradient} store and retrain on previous task data, but can pose privacy concerns. (\rmnum{2}) Regularization-based approaches~\cite{kirkpatrick2017overcoming, zenke2017continual} mitigate forgetting by imposing constraints that discourage significant changes in key weights. (\rmnum{3}) Architecture-based approaches~\cite{mallya2018packnet, wang2023rehearsal, liu2021l3doc, liu2023augmented, ning2024sparse} expand model capacity dynamically or isolate existing weights with task-specific parameters to minimize interference between new and old tasks.

\textbf{PEFT-Based Continual Learning for LLMs.}
LLMs with their vast number of parameters, create a significant computational burden for continual learning. 
To address this, methods like Learning to Prompt (L2P)~\cite{wang2022learning} use a pool of small, learnable prompts to guide pre-trained models through tasks, while DualPrompt~\cite{wang2022dualprompt} combines expert and general prompts for task-specific and task-agnostic instructions.
PEGP~\cite{qiao2024gradient} utilizes Singular Value Decomposition(SVD) of task data to control gradient updates through orthogonal gradient projection, ensuring that new parameters do not interfere with the established feature space. O-LoRA~\cite{wang2023orthogonal} incrementally updates new task parameters by constraining them to the orthogonal subspace of previous tasks while fixing the LoRA parameters learned from past tasks to minimize catastrophic forgetting. However, although O-LoRA attempts to update parameters within an orthogonal subspace, it struggles to prevent parameter collisions, which ultimately fails to resolve the forgetting problem. Our proposed N-LoRA not only achieves better orthogonality than O-LoRA but also reduces parameter collisions and delivers superior performance.

\section{Methodology}
\subsection{Continual Learning for LLMs}
In this paper, we focus on continual learning for LLMs, where the model adapts sequentially to new tasks presented in a streaming manner without access to previous task data.

\textbf{Orthogonality-Based CL for LLMs.} Continual learning in LLMs poses computational challenges due to their large parameter space. To mitigate this, existing studies have used PEFT methods and employed orthogonal gradient descent-based approaches to decouple parameter interdependence across different tasks.
Among these, O-LoRA introduces matrices $A \in \mathbb{R}^{d \times r}$ and, $B \in \mathbb{R}^{r \times k}$ based on the low-rank assumption~\cite{wang2023orthogonal}, updating the parameters as $W = W_0 + \Delta W = W_0 + AB$.
O-LoRA enforces orthogonality between the task subspaces by imposing constraints on $A_t$ and $A_i$, introducing an additional orthogonality loss:
\begin{equation}
L_{\text {orth}}\left(A_{t}\right)=\sum_{i=1}^{t-1}\left\|A_{i}^{T} A_{t}\right\|^{2}
\end{equation}
However, despite the enforced orthogonality constraints, the parameter subspaces used by different tasks may still overlap, preventing the achievement of full orthogonality (orthogonality loss fails to converge). Additionally, LoRA parameters within different subspaces may interact, leading to potential parameter collisions, which in turn degrade the performance across tasks.


\subsection{Orthogonality vs. Non-Collision}
We reveal that non-collision LoRA parameters provide a more effective solution to parameter interdependence and show that non-collision is a sufficient but not necessary condition for orthogonality (see Appendix \ref{sec:appendix: Theorem 1} for proof). The detailed analysis is as follows, 


\begin{definition}
\label{def:non-collision_matrices}
    For two parameter matrices $\Delta W_1$ and $\Delta W_2$ of the same dimensions, they are defined as non-collision if for every position $(a, b)$, it holds that $\Delta W_1[a,b] = 0$ or $\Delta W_2[a,b] = 0$.
\end{definition}

\begin{definition}
    For two parameter matrices $\Delta W_1$ and $\Delta W_2$ of the same dimensions, they are said to be orthogonal if their matrix product satisfies $\Delta W_1^\top \Delta W_2 = 0$, where $\Delta W_1^\top$ denotes the transpose of $\Delta W_1$.
\end{definition}

\begin{theorem}
    For two parameter matrices $\Delta W_1$ and $\Delta W_2$ of the same dimensions, non-collision is a sufficient but not necessary condition for orthogonality. 
    Specifically,
    \[
    \forall (a,b), \quad \Delta W_1[a,b] = 0 \vee \Delta W_2[a,b] = 0 
    \]
    \[
    \implies \Delta W_1^\top \Delta W_2 = 0,
    \]
\end{theorem}
Obviously, non-collision parameters inherently result in orthogonality. In other words, reducing parameter collisions is a more efficient way to achieve better task orthogonality. 



\begin{figure}[t]
  \includegraphics[width=\columnwidth]{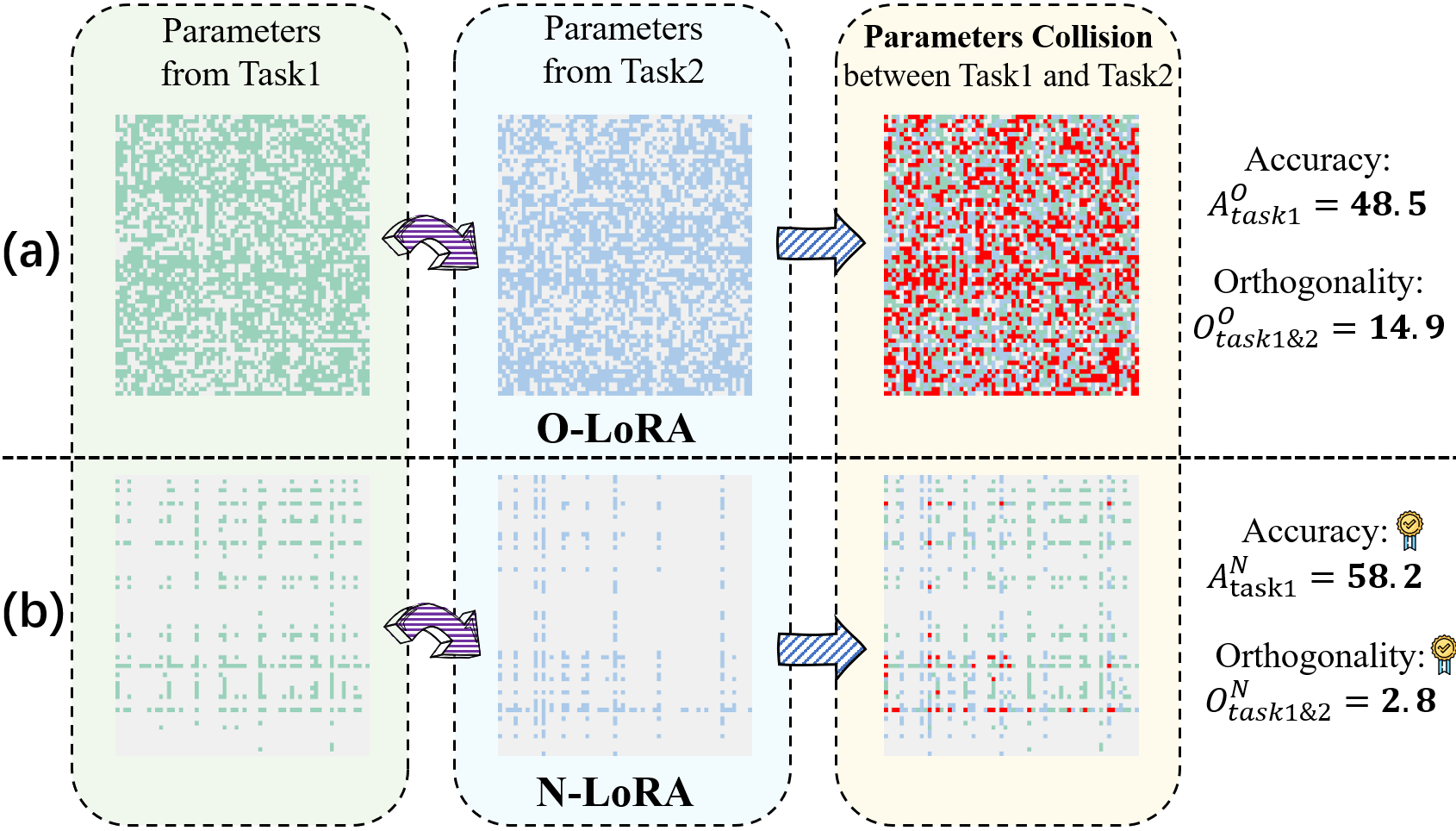}
  \caption{Parameter distribution of O-LoRA and N-LoRA. Green and blue represent parameters from Task 1 and Task 2, respectively; red indicates collisions. Lower orthogonality values signify better orthogonality. \textbf{(a) O-LoRA parameter distribution:} Despite the orthogonality constraint, significant parameter collisions occur, resulting in an accuracy of $\bm{48.5\%}$ and an orthogonality of $\bm{14.9}$. \textbf{(b) N-LoRA parameter distribution:} N-LoRA significantly reduces parameter collisions, achieving better performance with an accuracy of $\bm{58.2\%}$ and an orthogonality of $\bm{2.8}$.}
  \label{fig:2}
\end{figure}

\textbf{Visualization on Parameter Collisions.} Based on the above analysis, we aim to visualize the relationship between parameter orthogonality and non-collision. We conducted experiments on the T5-large model, comparing parameter distributions using O-LoRA (orthogonality-based) and N-LoRA (non-collision-based) in a continual learning setup for Task 1 and Task 2. In the visualizations, parameters near zero are shown in gray, non-zero parameters in green and blue, and collisions in red.
Figure \ref{fig:2} illustrates the parameter collisions under O-LoRA (top) and N-LoRA (bottom) methods. As shown in O-LoRA, despite the imposition of orthogonality constraints, significant parameter collisions (indicated by extensive red areas) were still observed, resulting in an accuracy of $\bm{48.5\%}$ and an orthogonality measure of $\bm{14.9}$. In contrast, N-LoRA effectively minimized parameter collisions (shown by small red areas), achieving an accuracy of $\bm{58.2\%}$ and an orthogonality measure of $\bm{2.8}$.
Thus, N-LoRA's low collision rate not only leads to higher accuracy but also results in better orthogonality.

\begin{figure}[t]
  \includegraphics[width=\columnwidth]{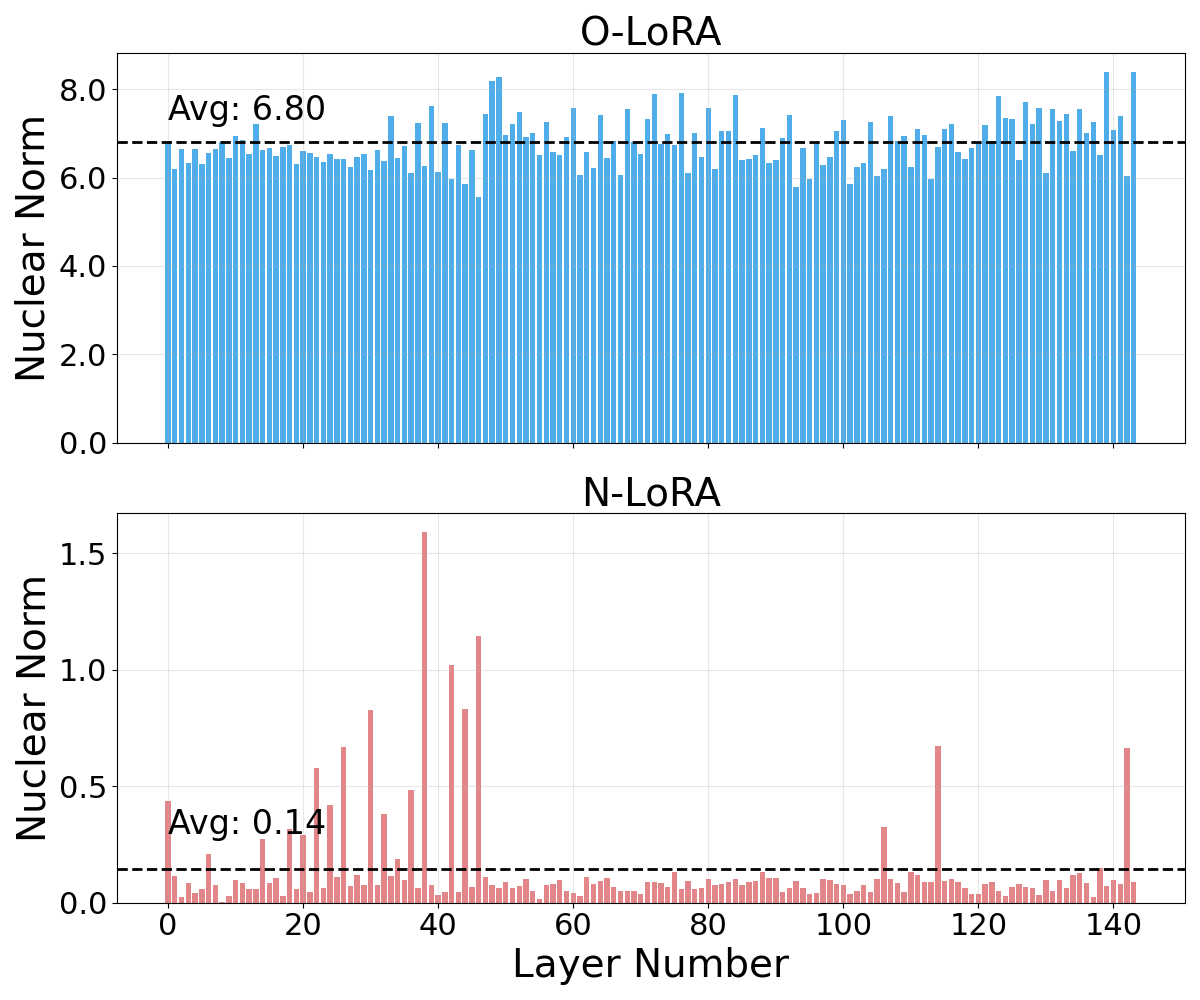}
  \caption{Nuclear norms of O-LoRA and N-LoRA. The top plot (blue) shows the nuclear norms for O-LoRA, indicating the subspace dimensionality used by each layer. The bottom plot (red) shows the nuclear norms for N-LoRA. The dashed lines in both plots represent the average nuclear norm across all layers.}
  \label{fig:4}
\end{figure}

\subsection{Non-collision Low-Rank Adaptation}
\label{section: N-lora}
Building on the motivation above, the core goal of tackling continual learning in LLMs is to minimize parameter collisions as much as possible, \textit{i.e.}, 
\begin{equation*}
    \min \sum_{\forall(a,b)} \mathbf{1}\{\Delta W_1[a,b] \neq 0 \land ... \land \Delta W_n[a,b] \neq 0\},
\end{equation*}
where $\mathbf{1}\{\cdot\}$ is the indicator function. To achieve this, we propose to sparsify each LoRA parameter $\Delta W_i$, \textit{i.e.}, 
\begin{equation}
    \max \sum_{\forall(a,b)} \mathbf{1}\{W_i[a,b]=0\}
\end{equation}
We attempt to sparsify individual LoRA parameters to minimize overall collisions. 
We show that \textit{\textbf{the collision rate will decrease quadratically with the sparsity rate}} in Appendix \ref{append:2}.
To further simplify, we introduce a sparsity constraint directly on $\Delta W_i$ while freezing the parameters from previous tasks. Specifically, for a sequence of tasks $T = \{t_1, t_2, \dots, t_n\}$, a corresponding sequence of task-specific matrices $\Delta W = \{\Delta W_1, \Delta W_2, \dots, \Delta W_n\}$ is obtained through continual fine-tuning using LoRA.  For the $i$-th task, the matrix $\Delta W_i = A_i B_i$ is subject to $\ell_1$ regularization:
\begin{equation}
    L_{\text{sparse}} = \lambda \ell_1(\Delta W_{i}) = \lambda \lVert \Delta W_{i} \rVert_1
\end{equation}
where $\lambda$ is the sparsity hyperparameter. For each new task, only the task-specific LoRA parameters are updated, while the pre-trained model and previously learned LoRA parameters remain fixed. Consequently, the overall loss function for the $i$-th task is defined as:
\begin{equation}
    L = L_{\text{task}} + L_{\text{sparse}} = L_{t_i} + \lambda \lVert \Delta W_{i} \rVert_1
\end{equation}

Additionally, since the parameters from previous tasks are not needed during training or inference, they can be merged with the pre-trained model’s parameters, effectively reducing memory usage and maintaining the model's efficiency.


\textbf{Discussion: N-LoRA can effectively reduce the dimension of task subspaces.}  
Beyond reducing collision rates, sparsity offers the additional advantage of effectively reducing the dimensionality of task subspaces. The nuclear norm, which provides an upper bound on the rank of a matrix~\cite{li2021concise}, corresponds to the number of subspaces utilized by each adapter in PEFT-based continual learning. Taking the first task of the Standard CL Benchmark as an example, we computed the nuclear norm of the LoRA weights inserted at each layer of the T5-large model for both O-LoRA and N-LoRA, as illustrated in Figure \ref{fig:4}. With the rank set to 8 (i.e., 8 subspaces per task), O-LoRA (top plot) nearly fully occupied these subspaces, averaging \textbf{6.80} across all layers. In contrast, N-LoRA, by leveraging sparsity, utilized an average of only \textbf{0.14} subspaces. This clearly demonstrates that N-LoRA significantly reduces the dimension of task subspaces. Furthermore, knowledge from multiple domains will be preserved in these extremely low-dimensional (non-collision) parameter subspaces, making it more difficult to forget previously seen data.


\section{Experiments}

\subsection{Experimental Setups}
\textbf{Datasets.}
To evaluate our method, we use the {\textit{Standard CL Benchmark}}~\cite{zhang2015character} with five text classification datasets and the {\textit{Large Number of Tasks Benchmark}}~\cite{razdaibiedina2023progressive} encompassing 15 datasets, including GLUE~\cite{wang2018glue}, SuperGLUE~\cite{wang2019superglue}, and IMDB~\cite{maas2011learning}. We follow the experimental setups of LFPT5~\cite{qin2021lfpt5} and O-LoRA~\cite{wang2023orthogonal}, using three different orders for each benchmark. Detailed task information and configurations are in Appendix \ref{sec:appendix: Datasets}, \ref{sec:appendix: Task Sequence and Instructions}, and \ref{sec:appendix: Implementation Details}.

\textbf{Metrics.}
Let $a_{i,j}$ be the test accuracy of the $i$-th task after training on the $j$-th task. $A_i$ denotes the LoRA $A$ matrix and $\Delta W_i =A_i B_i$ matrix, both after training on the $i$-th task. We evaluate the model using the following six metrics:\\
(\Rmnum{1}) \textbf{Average Accuracy (AA)}~\cite{lopez2017gradient}: The average accuracy of all tasks after training on the final task, defined as: 
\begin{equation}
    A_{T}=\frac{1}{T} \sum_{i=1}^{T} a_{i,T}
\end{equation}
(\Rmnum{2}) \textbf{Forgetting Rate (F.Ra)}~\cite{chaudhry2018riemannian}: This metric evaluates the amount of knowledge forgotten across the first $T-1$ tasks, as:
\begin{equation}
    F_{T} = \frac{1}{T-1} \sum_{t=1}^{T-1} \left( \max_{k=t}^{T-1} a_{k,t} - a_{T,t} \right)
\end{equation}
(\Rmnum{3}) \textbf{Orthogonal Overlap (OO)}~\cite{wang2023orthogonal}: Follow the O-LoRA, it quantifies the orthogonal overlap between the subspace of the $T$-th task and the subspaces of the previous $T-1$ tasks, calculated as:
\begin{equation}
    O_{T} = \sum_{i=1}^{t-1}\left\|A_{i}^{T} A_{t}\right\|^{2}
\end{equation}
(\Rmnum{4}) \textbf{Adapter Weight Orthogonality Magnitude (AWOM)}~\cite{bhardwaj2024sparse}: This metric measures the degree of orthogonality between adapter weights, defined as:
\begin{equation}
    AWOM_T = \sum_{i=1}^{T-1} \left\|\Delta W_{T}^\top \Delta W_{i} \right\|_2
\end{equation}

(\Rmnum{5}) \textbf{Generalized Sparsity Rate (GSR)}~\cite{hurley2009comparing}: This metric quantifies the sparsity of a matrix when none of its elements are zero. For detailed information and proof, see Appendix \ref{sec:appendix: Supplementary Explanation}. The formula is:
\begin{equation}
\begin{aligned}
    S_{q, p}^{*}(\Delta W_i) &= (mn)^{1/p - 1/q} \frac{\ell_q(\Delta W_i)}{\ell_p(\Delta W_i)},  (p > q > 0)
\end{aligned}
\end{equation}
where \( m \times n \) is the dimension of \( \Delta W_i \), and $\ell_p$ and $\ell_q$ are the respective norms. In our case, \( p = 2 \) and \( q = 1 \).

\begin{table*}[t]
    \renewcommand{\arraystretch}{1}
    \setlength{\tabcolsep}{7pt}
    \renewcommand{\arraystretch}{0.7}
    \centering
    \begin{tabular}{c | c c c c | c c c c}
    \toprule
        Benchmarks & \multicolumn{4}{c|}{Standard CL Benchmark} & \multicolumn{4}{c}{Large Number of Tasks}  \\ 
        Methods & Order-1 & Order-2 & Order-3 & avg & Order-4 & Order-5 & Order-6 &  avg \\ 
        \toprule
        ProgPrompt & 75.2 & 75.0 & 75.1 & 75.1 & 78.0 & 77.7 & 77.9 & 77.9 \\ 
        PerTaskFT & 70.0 & 70.0 & 70.0 & 70.0 & 78.1 & 78.1 & 78.1 & 78.1 \\
        MTL & 80.0 & 80.0 & 80.0 & 80.0 & 76.5 & 76.5 & 76.5 & 76.5 \\
        \toprule
        SeqFT & 18.9 & 24.9 & 41.7 & 28.5 & 7.4 & 7.4 & 7.5 & 7.4 \\
        SeqLoRA & 44.6 & 32.7 & 53.7 & 43.7 & 0.6 & 1.9 & 1.6 & 1.6 \\
        IncLoRA & 66.0 & 64.9 & 68.3 & 66.4 & 63.3 & 58.5 & 61.7 & 61.2 \\
        Replay & 55.2 & 56.9 & 61.3 & 57.8 & 55.0 & 54.6 & 53.1 & 54.2 \\
        EWC & 48.7 & 47.7 & 54.5 & 50.3 & 45.3 & 44.5 & 45.6 & 45.1 \\
        LwF & 54.4 & 53.1 & 49.6 & 52.3 & 50.1 & 43.1 & 47.4 & 46.9 \\
        L2P & 60.3 & 61.7 & 61.1 & 60.7 & 57.5 & 53.8 & 56.9 & 56.1 \\
        LFPT5 & 67.6 & 72.6 & 77.9 & 72.7 & 70.4 & 68.2 & 69.1 & 69.2 \\ 
        O-LoRA & 75.4 & 75.7 & 76.3 & 75.8 & 72.3 & 64.8 & 71.6 & 69.6 \\
        \textbf{N-LoRA} & $\bm{79.2}$ & $\bm{78.4}$ & $\bm{78.8}$ & $\bm{78.8}$ & $\bm{73.6}$ & $\bm{70.3}$ & $\bm{73.2}$ & $\bm{72.4}$ \\
        \toprule
    \end{tabular}
    \caption{Summary of results on two standard continual learning benchmarks using the T5-large model. The average accuracy after training on the final task is reported, with all results averaged over three independent runs.}
    \label{table: 1}
\end{table*}

\noindent(\Rmnum{6}) \textbf{Average Collision Rate (ACR)}\label{metric: CR}:  The concept of parameter collision was mentioned in ~\cite{kirkpatrick2017overcoming}. Here, we formally define the evaluation metric ACR, which quantifies the Collision Rate (CR) between parameters across tasks. It is defined as:
\begin{equation}
    ACR_T = \frac{2}{T(T-1)} \sum_{1 \leq i < j \leq T} CR_{i,j}
\end{equation}
where $CR_{i,j}$ is the collision rate between the $i$-th and $j$-th task' parameters, which is the fraction of elements that are non-zero at the same positions in both matrices, as defined in Definition~\ref{def:non-collision_matrices}:
\begin{equation}
    CR_{i,j} = \frac{\sum_{a,b} \mathbf{1}\left(\Delta W_i[a,b] \neq 0 \land \Delta W_j[a,b] \neq 0 \right)}{n \times m}
\end{equation}
\footnotetext{With the exception of the results for O-LoRA and N-LoRA, which were obtained as part of this work, all other data in Table\ref{table: 1} are based on those reported by~\cite{wang2023orthogonal}, using the same experimental setup.}

\textbf{Implementation Details}
Our experiments on the T5-large model were conducted on machines equipped with NVIDIA A6000 GPUs. For tasks in orders 1 to 3, we used a learning rate of 1e-3, a batch size of 32, a dropout rate of 0.1, and a weight decay rate of 0. The models were trained for 10 epochs with $\lambda = 0.4$. For tasks in order 4, we set the learning rate to 5e-4, the batch size to 32, the dropout rate to 0.1, and the weight decay rate to 0. The models were trained for 10 epochs, and $\lambda$ values were set as follows: (0.4, 0.4, 1.2, 0.4, 0.4, 0.4, 1.2, 1.2, 1.2, 0.4, 1.2, 1.2, 0.4, 0.4, 1.2). For tasks in order 5, we used a learning rate of 8e-4, a batch size of 32, a dropout rate of 0.1, and a weight decay rate of 0. The models were trained for 10 epochs with $\lambda = 0.55$. For tasks in order 6, we set the learning rate to 1e-3, the batch size to 32, the dropout rate to 0.1, and the weight decay rate to 0. The models were trained for 10 epochs with $\lambda = 1.8$. 

Details for the specific task sequences in each order can be found in Appendix~\ref{sec:appendix: Task Sequence and Instructions}. Details for experiments on the LLAMA-7B model can be found in Appendix~\ref{sec:appendix: Implementation Details}.

\textbf{Baselines.}
To demonstrate the effectiveness of N-LoRA, we follow the setup of O-LoRA ~\cite{wang2023orthogonal} and compare it against the following baseline methods:

\textit{(\rmnum{1}) Independent Training Methods}: \textbf{PerTaskFT} trains a separate model for each task, while \textbf{ProgPrompt}~\cite{razdaibiedina2023progressive} learns independent prompts for each task, which essentially results in training a distinct model for each task. 

\textit{(\rmnum{2}) Upper Bound}: \textbf{MTL} represents multi-task learning, where a single model is trained on all tasks simultaneously. This method serves as the upper bound for continual learning. 

\textit{(\rmnum{3}) Non-Continual Learning Methods}: \textbf{SeqFT}~\cite{de2019episodic} fine-tunes all model parameters across a sequence of tasks. \textbf{SeqLoRA} trains a single LoRA adapter across the task sequence while freezing the pretrained model parameters. \textbf{IncLoRA} trains a new LoRA module for each task. 

\textit{(\rmnum{4}) Continual Learning Methods}: \textbf{Replay} fine-tunes all model parameters while replaying samples from previous tasks using a memory buffer. \textbf{EWC}~\cite{kirkpatrick2017overcoming} applies a regularization loss to adjust all model parameters. \textbf{LwF}~\cite{li2017learning} preserves knowledge from previous tasks by recording their responses on new task data and using these responses as a regularization term. Both \textbf{L2P}~\cite{wang2022learning} and \textbf{LFPT5}~\cite{qin2021lfpt5} employ prompts, dynamically selecting or generating prompts to adapt to new tasks. \textbf{O-LoRA}~\cite{wang2023orthogonal} incrementally updates parameters for new tasks by constraining them to the orthogonal subspace of previous tasks while keeping the LoRA parameters learned from prior tasks fixed.

\subsection{Comparisons with SOTA methods}

Tables \ref{table: 1} and \ref{table: 2} present the performance comparison (Average Accuracy $A_T$) between N-LoRA and other baselines on different benchmarks. Following LFPT5~\cite{qin2021lfpt5} and O-LoRA~\cite{wang2023orthogonal}, we report the results from three independent runs using different task orders and models on the CL benchmark.

\textbf{Performance on Standard CL Benchmarks.}  
N-LoRA consistently outperformed all previous methods across all task orders by a significant margin on standard CL benchmarks. 
In particular, N-LoRA achieved over a $\bm{3.0\%}$ improvement compared to the previous SOTA method, O-LoRA. 
Additionally, N-LoRA surpassed independent training methods such as ProgPrompt and PerTaskFT, and its performance approached that of MTL, the upper bound for continual learning.
This indicates that N-LoRA is capable of efficiently leveraging knowledge from multiple tasks. 
As shown in Table \ref{table: 2}, N-LoRA also significantly outperformed O-LoRA on the larger LLaMA-7B model, demonstrating that our method remains effective even on larger-scale LLMs.
\begin{table}[t]
    \renewcommand{\arraystretch}{1}
    \setlength{\tabcolsep}{2pt}
    \renewcommand{\arraystretch}{0.8}
    \centering
    \begin{tabular}{c | c c c c}
    \toprule
        Benchmarks & Order-1 & Order-2 & Order-3 & avg\\ 
        \toprule
        O-LoRA & 76.8 & 75.7 & 75.7 & 76.1\\ 
        N-LoRA & $\bm{77.2}$ & $\bm{77.3}$ & $\bm{78.4}$ & $\bm{77.6}$  \\ 
        \toprule
    \end{tabular}
    \caption{Performance comparison between N-LoRA and O-LoRA on the larger LLaMA-7B model, reporting average accuracy across all task orders.  Results are averaged over three independent runs.}
    \label{table: 2}
\end{table}

\textbf{Performance on a Large Number of Tasks.}  
Continual learning with a large number of tasks presents a significantly more challenging benchmark. Among the evaluated methods, SeqLoRA exhibits the worst performance due to significant parameter collisions and overlaps in non-orthogonal spaces. Its metrics, including OO (3796.23), AWOM (1120.5), Sparsity Rate (0.682), and ACR (0.542), reflect severe inefficiencies, limiting its ability to maintain task-specific knowledge.

In contrast, as shown in Table \ref{table: 1}, N-LoRA surpasses all existing SOTA methods across task orders (4, 5, and 6), achieving an average performance improvement of $\bm{2.8\%}$. By reducing parameter collisions and optimizing subspace allocation, N-LoRA delivers significantly better metrics, outperforming SeqLoRA by factors of approximately 1k, 8k, 85, and 71, respectively. These results underscore N-LoRA’s superior capacity to handle numerous tasks while preserving task-specific knowledge. Further comparisons with O-LoRA are detailed in Appendix~\ref{appendix:comparison_O-lora}.

\begin{figure}[t]
 \centering
  \includegraphics[width=\columnwidth]{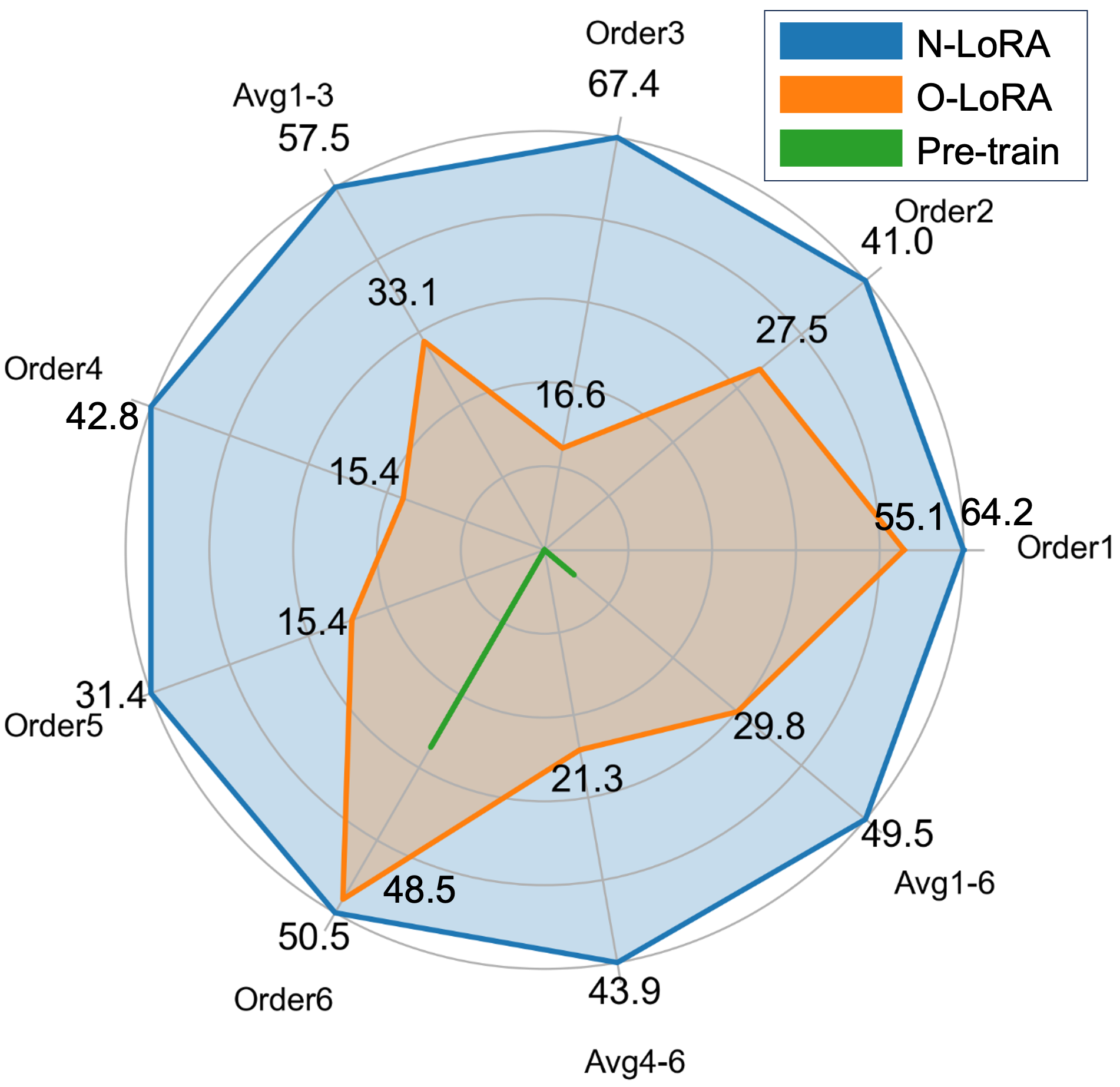}
  \caption{Performance comparison on unseen tasks on Average Accuracy metric. N-LoRA consistently outperforms O-LoRA across all task order settings. The pre-trained model performs poorly, with accuracy close to random on unseen tasks.}
  \label{fig:6}
\end{figure}


\subsection{Enhancing Generalization of LLMs}
This subsection explores how N-LoRA affects the generalization ability of LLMs, evaluated on the Standard CL Benchmark and Large Number of Tasks Benchmark. Using both N-LoRA and O-LoRA (current SOTA method), we conduct continual learning across the first $T-1$ tasks, followed by testing on the unseen $t$-th task. For comparison, we also evaluate pre-trained LLMs directly on the unseen task.

As shown in Figure \ref{fig:6}, the pre-trained models perform nearly at random, achieving close to zero accuracy on the unseen task, while N-LoRA achieves an average performance of $\bm{49.54\%}$. Moreover, N-LoRA delivers an average performance improvement of $\bm{+19.78\%}$ over O-LoRA, consistently outperforming it across all orders (1-6). These findings underscore the effectiveness of N-LoRA in enhancing strong generalization to unseen tasks.

\subsection{N-LoRA vs O-LoRA: Reducing Forgetting}
In this subsection, we conduct an in-depth analysis of how N-LoRA reduces forgetting from multiple perspectives, providing a comprehensive comparison with O-LoRA, the current SOTA method.

\begin{table}[t]
    \renewcommand{\arraystretch}{1}
    \setlength{\tabcolsep}{1.8pt}
    \centering
    \renewcommand{\arraystretch}{0.8}
    \begin{tabular}{ c | c c | c c}
    \toprule
        Metrics & \multicolumn{2}{c|}{OO$(\downarrow)$} & \multicolumn{2}{c}{AWOM$(\downarrow)$}\\ 
         Order & O-LoRA & N-LoRA & O-LoRA & N-LoRA\\ 
        \toprule
         1 & 21.00 & $\bm{3.80}$ & 55.42 & $\bm{0.13}$ \\ 
         2 & 20.50 & $\bm{4.16}$ & 126.25 & $\bm{0.17}$ \\ 
        3 & 20.13 & $\bm{3.42}$ & 118.75 & $\bm{0.15}$ \\ 
        4 & 30.75 & $\bm{13.94}$ & 14.40 & $\bm{0.19}$ \\ 
        5 & 31.38 & $\bm{11.06}$ & 14.40 & $\bm{0.19}$ \\ 
        6 & 34.50 & $\bm{2.47}$ & 6.57 & $\bm{0.03}$ \\ 
        avg & 26.38 & $\bm{6.47}$ & 55.96 & $\bm{0.14}$ \\
        \toprule
    \end{tabular}
    \caption{Performance comparison between O-LoRA and N-LoRA on Orthogonal Overlap (OO) and Adapter Weight Orthogonality Magnitude (AWOM) metrics. Lower values indicate better orthogonality.}
    \label{table: 3}
\end{table}

\begin{table*}[t]
    \renewcommand{\arraystretch}{1}
    \setlength{\tabcolsep}{5pt}
    \renewcommand{\arraystretch}{0.7}
    \centering
    \begin{tabular}{c | c c c c | c c c c}
    \toprule
        Benchmarks & \multicolumn{4}{c|}{Standard CL Benchmark} & \multicolumn{4}{c}{Large Number of Tasks}  \\ 
        Methods & Order-1 & Order-2 & Order-3 & avg & Order-4 & Order-5 & Order-6 &  avg \\ 
        \toprule
        O-LoRA & 75.4 & 75.7 & 76.3 & 75.8 & 72.3 & 64.8 & 71.6 & 69.6 \\
        \textbf{O-LoRA + N-LoRA} & $\bm{79.7}$ & $\bm{79.8}$ & $\bm{79.1}$ & $\bm{79.5}$ & $\bm{72.9}$ & $\bm{67.2}$ & $\bm{74.0}$ & $\bm{71.4}$  \\
        \toprule
    \end{tabular}
    \caption{Performance comparison of O-LoRA and O-LoRA+N-LoRA on two benchmarks. \textbf{O-LoRA+N-LoRA} represents the integration of N-LoRA into O-LoRA, enhancing the performance of O-LoRA across all task orders.}
    \label{table: 4}
\end{table*}

\textbf{Higher Orthogonality.} We adopt Orthogonal Overlap (OO)~\cite{wang2023orthogonal} and Adapter Weight Orthogonality Magnitude (AWOM)~\cite{bhardwaj2024sparse} as metrics to assess orthogonality. As shown in Table \ref{table: 3}, N-LoRA surpasses O-LoRA across both metrics, achieving \textbf{$\bm{\times4.07}$ times} and \textbf{$\bm{\times388.3}$ times} improvements, respectively. These results further demonstrate that the low collision rate induced by sparsity constraints can achieve stronger orthogonality compared to orthogonality constraints.
Notably, although O-LoRA directly optimizes orthogonality using Orthogonal Overlap (OO) as its loss function, its actual OO is still lower than that of N-LoRA. This can likely be attributed to significant overlap in the task subspaces, leading to unavoidable conflicts, despite the orthogonal constraint. In contrast, as discussed in Section \ref{section: N-lora}, N-LoRA reduces the dimensionality of task subspaces, decreasing the collison rate and consequently enhancing orthogonality.

\begin{figure}[t]
  \includegraphics[width=\columnwidth]{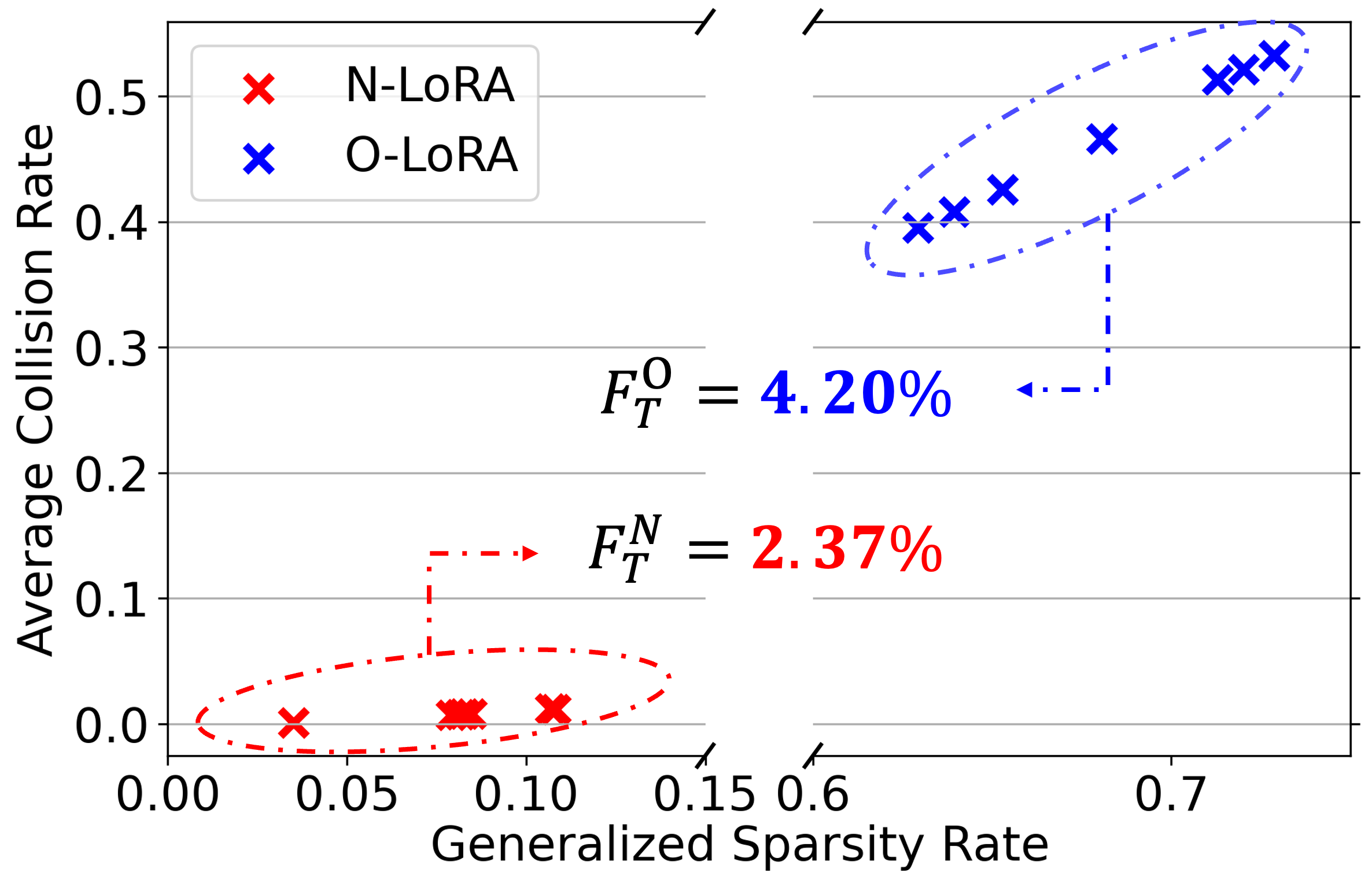}
  \caption{The relationship between sparsity, collision rate, and forgetting. The x-axis and y-axis represent the Generalized Sparsity and Average Collision Rate, respectively. Data points for N-LoRA (red) are clustered in the lower-left corner, while O-LoRA (blue) data points are concentrated in the upper-right corner. The dashed ellipses indicate the average forgetting rates for each method.}
  \label{fig:5}
\end{figure}

\textbf{Lower Sparsity Leads to Fewer Collisions and Reduced Forgetting.}  
Figure \ref{fig:5} illustrates the relationship between sparsity, collision rate, and forgetting rate for N-LoRA and O-LoRA. It can be observed that the data points for N-LoRA are clustered in the lower-left corner of the plot, with an average forgetting rate of $\bm{2.37\%}$. In contrast, the data points for O-LoRA are in the upper-right corner, with an average forgetting rate of $\bm{4.20\%}$. This indicates that across varying benchmarks and  orders, lower sparsity leads to fewer collisions and consequently reduces forgetting. The sparsity constraint in N-LoRA effectively minimizes collision rates, directly contributing to lower forgetting. Conversely, even though O-LoRA applies an orthogonality constraint on subspaces, it still encounters a high collision rate, resulting in greater forgetting.

\textbf{Plug-and-Play Integration.}  
Beyond its standalone effectiveness, N-LoRA can be seamlessly integrated into existing PEFT-based continual learning methods, acting as a versatile plug-and-play solution. 
As shown in Table \ref{table: 4}, integrating N-LoRA into O-LoRA, resulting in "O-LoRA + N-LoRA", led to significant performance improvements across all task orders on the Standard CL Benchmark, with increases of $\bm{4.3\%}$, $\bm{4.1\%}$, and $\bm{3.2\%}$ for Order-1, Order-2, and Order-3, respectively.
These results not only surpass independent training methods but also approach the upper performance limit set by MTL, which is 80$\%$.
Additionally, "O-LoRA + N-LoRA" outperformed O-LoRA on all task orders in the Large Number of Tasks Benchmark, further demonstrating its robustness in managing a wide range of tasks and consistently enhancing continual learning performance.

\begin{figure}[t]
  \includegraphics[width=\columnwidth]{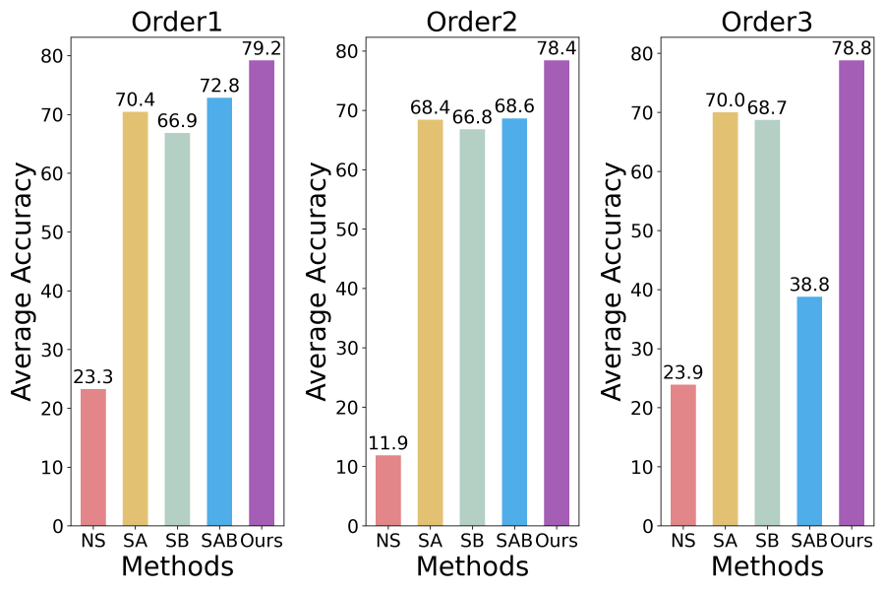}
  \caption{Ablation study comparing different sparsity constraint methods. NS refers to no sparsity constraint, while SA and SB apply sparsity constraints only to LoRA matrices $A$ and $B$, respectively. SAB applies independent sparsity constraints to both matrices $A$ and $B$. Ours represents the N-LoRA method, which significantly outperforms all other methods.}
  \label{fig:7}
\end{figure}

\subsection{Ablation Studies}

In this subsection, we perform ablation studies to investigate the impact of sparsity constraints and the various ways of applying them. As illustrated in Figure \ref{fig:7}, the model experiences severe forgetting when no sparsity constraint is applied (NS method). 
Applying sparsity to only LoRA matrix $A$ or matrix $B$ (SA and SB methods) can partially enhance parameter updates, resulting in some performance gains. However, since $\Delta W$ is the product of both $A$ and $B$, constraining only one matrix still permits considerable redundancy in the overall update.
Simultaneously applying constraints to both $A$ and $B$ (SAB method) also fails to achieve optimal results. This is because the actual update during forward propagation is $\Delta W = AB$, and constraining the matrices independently hampers effective gradient flow, thereby impeding global optimization.
The experimental results clearly show that only the N-LoRA approach (Ours) effectively reduces parameter collisions and substantially improves performance.

\section{Conclusion}
In this paper, we theoretically and empirically reveal that constructing non-collision parameters is a more critical factor in overcoming continual learning challenges. Building on this insight, we propose N-LoRA, a simple yet effective approach that leverages sparsity and low collision rates to improve continual learning. Extensive experiments on continual learning benchmarks demonstrate that, compared to SOTA methods, N-LoRA not only achieves better performance, higher task orthogonality, and lower parameter collision rates through sparsity, but also effectively reduces the dimensionality of task subspaces.


\section{Limitations}
Although N-LoRA demonstrates superior performance across multiple metrics in continual learning benchmarks, several limitations remain. In scenarios with a significantly larger number of tasks (e.g., hundreds or thousands), the parameter space can still become saturated, despite the imposed sparsity constraint. However, N-LoRA utilizes fewer subspaces than conventional LoRA, allowing it to reach saturation more slowly. Its scalability and effectiveness in such scenarios, as well as in highly dynamic real-world continual learning tasks, require further investigation.

The challenges of managing a large number of tasks and addressing concept drift remain open in the CL-LLM field. 
By minimizing parameter collisions and optimizing subspace allocation, N-LoRA demonstrates improved performance over longer sequences and more tasks. Effective subspace allocation also mitigates bias accumulation during concept drift, reducing abrupt performance drops and promoting more gradual declines. 
In future work, we will explore N-LoRA's scalability for larger tasks, longer sequences, and dynamic environments with concept drift.

\section*{Acknowledgments}
This work was supported in part by the Natural Science Foundation of China (No. 62202014, 62332002, 62425101) and the China Postdoctoral Science Foundation (No. BX20240013, 2024M760113). This work was also supported by DAMO Academy through DAMO Academy Innovative Research Program.


\clearpage

\appendix

\section{Appendix}
\label{sec:appendix}

\subsection{Proof of Theorem 1}
\label{sec:appendix: Theorem 1}
In this section, we provide the detailed proof for the theorem presented in Section 3.2, which states that non-collision is a sufficient but not necessary condition for orthogonality.

\begin{proof}
    \textbf{(1) Sufficiency:} 
    Assume that $\Delta W_1$ and $\Delta W_2$ are non-collision. By definition, for any position $(a,b)$, either $\Delta W_1[a,b] = 0$ or $\Delta W_2[a,b] = 0$. Consider any element $(k,l)$ of the matrix product $\Delta W_1^\top \Delta W_2$:
    $$
    (\Delta W_1^\top \Delta W_2)_{kl} = \sum_a \Delta W_1[a,k] \cdot \Delta W_2[a,l]
    $$

    Since $\Delta W_1[a,k]$ and $\Delta W_2[a,l]$ cannot both be non-zero simultaneously, each term in the sum is zero. Thus, the entire sum is zero, implying:
    $$
    \Delta W_1^\top \Delta W_2 = 0
    $$
    Therefore, $\Delta W_1$ and $\Delta W_2$ are orthogonal.
    
    \textbf{(2) Non-necessity:} Assume that $\Delta W_1$ and $\Delta W_2$ are orthogonal, i.e.,
    $$
    \Delta W_1^\top \Delta W_2 = 0.
    $$
    This implies that for any element $(k,l)$ of the matrix product:
    $$
    \sum_a \Delta W_1[a,k] \cdot \Delta W_2[a,l] = 0.
    $$
    However, orthogonality only requires that the sum of these products equals zero. It does not imply that each individual product $\Delta W_1[a,k] \cdot \Delta W_2[a,l] = 0$ for all $(a,k)$. As long as the sum is zero, some terms can be non-zero, leading to possible collisions. 
\end{proof}

\subsection{Relationship Between Collision Rate and Sparsity}
\label{append:2}
\begin{theorem}
    Consider two $m \times n$ parameter matrices, $\Delta W_i$ and $\Delta W_j$, with sparsity rates $s_i$ and $s_j$, respectively. If the non-zero elements in each matrix are independently and randomly distributed, the collision rate $CR_{i,j}$ between these matrices is given by:

    \begin{equation} 
    CR_{i,j} = s_i \times s_j 
    \end{equation}
\end{theorem}
Thus, as the sparsity rate increases, the collision rate decreases quadratically.

\begin{proof}

\textbf{Assumption:} The positions of non-zero elements in matrices $\Delta W_i$ and $\Delta W_j$ are independent and randomly distributed.

At any position $(a, b)$, the probability that both $\Delta W_i[a,b]$ and $\Delta W_j[a,b]$ are non-zero is $s_i \times s_j$. Since the elements are independently distributed, the expected total number of collisions in the matrix is the product of this collision probability and the total number of elements $(m \times n)$:
\begin{equation} 
E_{\text{collisions}} = (m \times n) \times (s_i \times s_j) 
\end{equation}

The collision rate \( CR_{i,j} \) (as defined in section \ref{metric: CR}) is obtained by dividing the expected total number of collisions by the total number of elements:
\begin{equation} 
CR_{i,j} = \frac{E_{\text{collisions}}}{m \times n} = \frac{(m \times n) \times s_i \times s_j}{m \times n} = s_i \times s_j 
\end{equation}

Therefore, the collision rate $CR_{i,j}$ is directly proportional to the product of the sparsity rates $s_i$ and $s_j$. Specifically, when $s_i = s_j = s$, the collision rate is:
\begin{equation} 
CR_{i,j} = s^2 
\end{equation}
\end{proof}
This confirms that the collision rate decreases as a quadratic function of the sparsity rate.

\subsection{Datasets}
\label{sec:appendix: Datasets}
\begin{table*}[ht]
\centering
\begin{tabular}{c|l l l l l }
\hline
\textbf{No.} & \textbf{Dataset name} & \textbf{Category} & \textbf{Task} & \textbf{Domain} & \textbf{Metric} \\
\hline
1 & Yelp & CL Benchmark & sentiment analysis & Yelp reviews & accuracy \\
2 & Amazon & CL Benchmark & sentiment analysis & Amazon reviews & accuracy \\
3 & DBpedia & CL Benchmark & topic classification & Wikipedia & accuracy \\
4 & Yahoo & CL Benchmark & topic classification & Yahoo Q\&A & accuracy \\
5 & AG News & CL Benchmark & topic classification & news & accuracy \\
6 & MNLI & GLUE & NLI & various & accuracy \\
7 & QQP & GLUE & paragraph detection & Quora & accuracy \\
8 & RTE & GLUE & NLI & news, Wikipedia & accuracy \\
9 & SST-2 & GLUE & sentiment analysis & movie reviews & accuracy \\
10 & WiC & SuperGLUE & word sense disambiguation & lexical databases & accuracy \\
11 & CB & SuperGLUE & NLI & various & accuracy \\
12 & COPA & SuperGLUE & QA & blogs, encyclopedia & accuracy \\
13 & BoolQA & SuperGLUE & boolean QA & Wikipedia & accuracy \\
14 & MultiRC & SuperGLUE & QA & various & accuracy \\
15 & IMDB & SuperGLUE & sentiment analysis & movie reviews & accuracy \\
\hline
\end{tabular}
\caption{Datasets and Their Tasks, Categories, Domains, and Metrics}
\label{apptable:2}
\end{table*}

In our experiments, we utilized 15 datasets, as detailed in Table \ref{apptable:2}, covering multiple evaluation metrics relevant to continual learning (CL) tasks. Specifically, the datasets were sourced from well-established benchmarks such as CL benchmark~\cite{zhang2015character}, GLUE ~\cite{wang2018glue}, and SuperGLUE~\cite{wang2019superglue}. In addition, we incorporated the IMDB movie review dataset~\cite{maas2011learning}. These datasets span diverse tasks, including natural language inference (NLI), sentiment classification (SC), and topic classification (TC), ensuring a robust evaluation of the model's generalization across multiple tasks.

\subsection{Task Sequence and Instructions}
\label{sec:appendix: Task Sequence and Instructions}

\begin{table*}[ht]
\centering
\begin{tabular}{c p{10cm} }
\hline
\textbf{Task} & \textbf{Prompts} \\
\hline
NLI & What is the logical relationship between the "sentence 1" and the "sentence 2"? Choose one from the option. \\
\hline
QQP & Whether the "first sentence" and the "second sentence" have the same meaning? Choose one from the option. \\
\hline
SC & What is the sentiment of the following paragraph? Choose one from the option. \\
\hline
TC & What is the topic of the following paragraph? Choose one from the option. \\
\hline
BoolQA & According to the following passage, is the question true or false? Choose one from the option. \\
\hline
MultiRC & According to the following passage and question, is the candidate answer true or false? Choose one from the option. \\
\hline
WiC & Given a word and two sentences, whether the word is used with the same sense in both sentence? Choose one from the option. \\
\hline
\end{tabular}
\caption{Task Descriptions and Corresponding Prompts}
\label{apptable:3}
\end{table*}

\begin{table*}[ht]
\centering
\begin{tabular}{c c p{10cm} }
\hline
\textbf{Order} & \textbf{Model} & \textbf{Task Sequence} \\
\hline
1 & T5, LLaMA & dbpedia → amazon → yahoo → ag \\
2 & T5, LLaMA & dbpedia → amazon → ag → yahoo \\
3 & T5, LLaMA & yahoo → amazon → ag → dbpedia \\
\hline
4 & T5 & mnli → cb → wic → copa → qqp → boolqa → rte → imdb → yelp → amazon → sst-2 → dbpedia → ag → multirc → yahoo \\
5 & T5 & multirc → boolqa → wic → mnli → cb → copa → qqp → rte → imdb → sst-2 → dbpedia → ag → yelp → amazon → yahoo \\
6 & T5 & yelp → amazon → mnli → cb → copa → qqp → rte → imdb → sst-2 → dbpedia → ag → yahoo → multirc → boolqa → wic \\
\hline
\end{tabular}
\caption{Task Sequences for Different Models}
\label{apptable:4}
\end{table*}

We report the task sequences used for CL experiments on the T5 and LLAMA models, as shown in Table \ref{apptable:4}. Additionally, Table \ref{apptable:3} provides prompts for different tasks. NLI represents natural language inference tasks, including MNLI, RTE, and CB. SC refers to sentiment classification tasks, including Amazon, Yelp, SST-2, and IMDB, while TC denotes topic classification tasks, such as AG News, Dbpedia, and Yahoo.

\subsection{Detailed Comparison between N-LoRA and O-LoRA}
\label{appendix:comparison_O-lora}

\begin{table*}[t]
    \renewcommand{\arraystretch}{1}
    \setlength{\tabcolsep}{0.1pt}
    \renewcommand{\arraystretch}{0.7}
    \centering
    \begin{tabular}{c | c c c c | c c c c}
    \toprule
        Benchmarks & \multicolumn{4}{c|}{Standard CL Benchmark} & \multicolumn{4}{c}{Large Number of Tasks}  \\ 
        Methods & Order-1 & Order-2 & Order-3 & avg & Order-4 & Order-5 & Order-6 & avg \\ 
        \toprule
        O-LoRA & $75.03\textsuperscript{\tiny{$\pm0.82$}}$ & $76.12\textsuperscript{\tiny{$\pm0.37$}}$ & $76.58\textsuperscript{\tiny{$\pm0.76$}}$ & $75.91\textsuperscript{\tiny{$\pm0.65$}}$ & $72.39\textsuperscript{\tiny{$\pm1.30$}}$ & $64.55\textsuperscript{\tiny{$\pm3.87$}}$ & $71.03\textsuperscript{\tiny{$\pm3.24$}}$ & $69.32\textsuperscript{\tiny{$\pm3.42$}}$ \\
        \textbf{N-LoRA} & $\bm{79.14\textsuperscript{\tiny{$\pm0.41$}}}$ & $\bm{78.29\textsuperscript{\tiny{$\pm0.39$}}}$ & $\bm{77.91\textsuperscript{\tiny{$\pm1.93$}}}$ & $\bm{78.45}\textsuperscript{\tiny{$\pm0.51$}}$ & $\bm{73.85\textsuperscript{\tiny{$\pm0.53$}}}$ & $\bm{70.09\textsuperscript{\tiny{$\pm0.93$}}}$ & $\bm{73.07\textsuperscript{\tiny{$\pm1.42$}}}$ & $\bm{72.34}\textsuperscript{\tiny{$\pm1.62$}}$ \\
    \toprule
    \end{tabular}
    \caption{Detailed performance comparison between N-LoRA and O-LoRA on two standard continual learning benchmarks using the T5-large model. Results are reported as the mean and standard deviation (in the superscript) over five independent runs.}
    \label{table:comparison}
\end{table*}

To ensure a statistically significant comparison between O-LoRA and N-LoRA, we retrained and evaluated each experimental group five times as shown in Table\ref{table:comparison}. N-LoRA demonstrated superior performance over O-LoRA in terms of both mean and standard deviation. Additionally, N-LoRA demonstrates greater stability than O-LoRA, as evidenced by the smaller standard deviation across all task orders.

\subsection{Implementation Details}
\label{sec:appendix: Implementation Details}

For the experiments on the LLAMA-7B model, we used the following settings:  
For the tasks in order 1, we set the learning rate to (1e-3, 1e-4, 1e-4, 1e-4), the batch size to 4, the dropout rate to 0.1, and the weight decay rate to 0. The models were trained for (1, 10, 1, 1) epochs with $\lambda = 0.46$.  
For the tasks in order 2, the learning rate was set to (1e-4, 5e-5, 5e-5, 5e-5), the batch size to 4, the dropout rate to 0.1, and the weight decay rate to 0. The models were trained for (1, 5, 1, 1) epochs with $\lambda = 0.38$.  
For the tasks in order 3, we used a learning rate of 1e-4, a batch size of 4, a dropout rate of 0.1, and a weight decay rate of 0. The models were trained for (5, 10, 1, 1) epochs with $\lambda = 0.4$.






\subsection{Supplementary Explanation For Generalized Sparsity Rate}
\label{sec:appendix: Supplementary Explanation}

Narrowly speaking, "sparsity" refers to the presence of a large number of zero elements in data, with the simplest metric being the proportion of zeros. However, in this paper, the parameter matrix $\Delta W$ is derived from the product of LoRA matrices $B$ and $A$, making it unlikely for individual parameters to be exactly zero (as this would require the dot product of each row in $B$ and each column in $A$ to equal zero). 

The generalized sparsity measure $S(x)$ should satisfy a series of ideal properties to ensure its theoretical validity and consistency with logical reasoning.
Therefore, following the work of ~\cite{hurley2009comparing}, we adopt a broader definition of sparsity, the sparsity measure $S(x)$ is considered to meet the following six key properties:

\begin{property}
\textbf{Robin Hood Property (D1)}  When energy is transferred from a larger element to a smaller element in a vector, the sparsity should decrease. Assume in the vector $x$, $x_i > x_j$, and we transfer an amount $\alpha \ (0 < \alpha < (x_i - x_j)/2)$ from $x_i$ to $x_j$, resulting in a new vector $x'$. Then, the following should hold:
$$
S(x') < S(x)
$$
\end{property}
This property reflects that a more evenly distributed vector is less sparse.

\begin{property}
\textbf{Dilation Invariance (D2)}  For any positive scalar $\alpha > 0$, the sparsity measure should satisfy:
$$S(\alpha x) = S(x)$$
\end{property}
This implies that sparsity is a relative property, and multiplying all elements of the vector by a positive constant does not change its sparsity.

\begin{property}
\textbf{Rising Tide Property (D3)} When a positive scalar $\alpha$ is added to all elements of the vector, the sparsity should decrease. Specifically:
$$
S(x + \alpha) > S(x), \ \ \alpha > 0
$$
\end{property}
As all elements increase and approach non-zero values, the sparsity naturally decreases.

\begin{property}
\textbf{Cloning Invariance (D4)}  If the vector $x$ is replicated (cloned) one or more times, its sparsity should remain unchanged. For the cloned vector $x' = [x, x]$, the following should hold:
$$S(x') = S(x)$$
\end{property}
This property reflects that duplicating elements in the vector does not affect its sparsity.

\begin{property}
\textbf{Dominance by Maximum Value (P1)} When one element $x_i$ in the vector becomes sufficiently large, the sparsity should be dominated by that element. Specifically, as $x_i \to \infty$:
$$
S(x) \to \text{max}
$$
\end{property}
This property indicates that when one element is overwhelmingly large, the sparsity of the vector is almost entirely determined by that element.

\begin{property}
\textbf{Effect of Adding Zeroes (P2)}  When zero elements are added to the vector, its sparsity should increase. If we add a zero to the vector, resulting in $x' = [x, 0]$, then:
$$
S(x') > S(x)
$$
\end{property}
Adding zero elements increases the sparsity of the vector.

For a matrix $\Delta W_i$ of dimensions $m \times n$, we define the Generalized Sparsity Rate (GSR) as:
\begin{equation}
S_{q, p}^{*}(\Delta W_i) = (mn)^{1/p - 1/q} \frac{\ell_q(\Delta W_i)}{\ell_p(\Delta W_i)}, \quad (p > q > 0)
\end{equation}

where $\ell_p$ and $\ell_q$ are the respective matrix norms, and $m \times n$ is the dimension of the matrix.

In this paper, we specifically use $p = 2$ and $q = 1$, which gives the following sparsity measure\cite{kexuefm-9595}:
\begin{equation}
S_{1,2}^{*}(\Delta W_i) = \frac{\sqrt{mn} \cdot \ell_1(\Delta W_i)}{\ell_2(\Delta W_i)}
\end{equation}

where:
\begin{itemize}
    \item $\ell_1(\Delta W_i) = \sum_{a=1}^{m} \sum_{b=1}^{n} |\Delta W_i[a, b]|$ denotes the $\ell_1$ norm, which is the sum of the absolute values of all elements in the matrix.
    \item $ \ell_2(\Delta W_i) = \left( \sum_{i=a}^{m} \sum_{j=b}^{n} |\Delta W_i[a,b]|^2 \right)^{1/2}$ denotes the $\ell_2$ (Frobenius) norm of the matrix.
\end{itemize}

The sparsity measure $S_{1,2}^{*}(\Delta W_i)$ satisfies the six properties outlined above, ensuring its theoretical soundness and consistency. Specifically, it adheres to the following:

\begin{theorem}
The sparsity measure $S_{1,2}^{*}(\Delta W_i)$ defined for a matrix satisfies the following six properties:

1. Robin Hood Property (D1)

2. Dilation Invariance (D2)

3. Rising Tide Property (D3)

4. Cloning Invariance (D4)

5. Dominance by Maximum Value (P1)

6. Effect of Adding Zeroes (P2)
\end{theorem}

Next, we will demonstrate that the proposed formula for Generalized Sparsity Ratio (GSR) satisfies the properties outlined above.

\subsubsection{Proof for Robin Hood Property (D1)}
    \textbf{Statement:} Transferring a value from a larger element to a smaller element in the matrix reduces the overall sparsity (i.e., \(S^*_{1,2}(\Delta W'_i) > S^*_{1,2}(\Delta W_i)\)).

\begin{proof}
    Let \(\Delta W_i\) be an \(m \times n\) matrix, and suppose \(\Delta W_i[a_1, b_1] > \Delta W_i[a_2, b_2]\). Transfer an amount \(\alpha\) (where \(0 < \alpha < \frac{\Delta W_i[a_1,b_1] - \Delta W_i[a_2, b_2]}{2}\)) from \(\Delta W_i[a_1, b_1]\) to \(\Delta W_i[a_2, b_2]\), resulting in a new matrix \(\Delta W'_i\):
    \begin{align}
        \Delta W'_i[a_1, b_1] &= \Delta W_i[a_1, b_1] - \alpha, \\
        \Delta W'_i[a_2, b_2] &= \Delta W_i[a_2, b_2] + \alpha, \\
        \Delta W'_i[a, b] &= \Delta W_i[a, b],  \text{for all other } (a, b).
    \end{align}
    Compute \(||\Delta W'_i||_1\):
    Since the total sum remains unchanged:
    \begin{equation}
        \Vert \Delta W'_i \rVert_1 = \Vert \Delta W_i \rVert_1.
    \end{equation}
    Compute \(||\Delta W'_i||_2^2\):

\begin{align}
    &\Vert \Delta W'_i \rVert_2^2 = \Vert \Delta W_i \rVert_2^2 \\
    &\quad - (\Delta W_i[a_1, b_1]^2 - \Delta W_i[a_2, b_2]^2) \\
    &\quad + (\Delta W'_i[a_1, b_1]^2 - \Delta W'_i[a_2, b_2]^2) \\
    &= \Vert \Delta W_i \rVert_2^2 - 2\alpha(\Delta W_i[a_1, b_1] - \Delta W_i[a_2, b_2]) + 2\alpha^2.
\end{align}

Since \(\Delta W_i[a_1, b_1] > \Delta W_i[a_2, b_2]\), the term \(-2\alpha(\Delta W_i[a_1, b_1] - \Delta W_i[a_2, b_2])\) is negative, leading to:

\begin{equation}
    \Vert \Delta W'_i \rVert_2^2 < \Vert \Delta W_i \rVert_2^2 \implies \Vert \Delta W'_i \rVert_2 < \Vert \Delta W_i \rVert_2.
\end{equation}

\textbf{Compute \(S^*_{1,2}(\Delta W'_i)\):}

\begin{align}
    S_{1,2}^*(\Delta W'_i) &= \frac{\sqrt{mn} \cdot \Vert \Delta W'_i \rVert_1}{\Vert \Delta W'_i \rVert_2} \\
    & = \frac{\sqrt{mn} \cdot \Vert \Delta W_i \rVert_1}{\Vert \Delta W'_i \rVert_2} \\
    & > \frac{\sqrt{mn} \cdot \Vert \Delta W_i \rVert_1}{\Vert \Delta W_i \rVert_2} = S_{1,2}^*(\Delta W_i).
\end{align}

Therefore, \(S^*_{1,2}(\Delta W'_i) > S^*_{1,2}(\Delta W_i)\), satisfying the Robin Hood Property.

\end{proof}

\subsubsection{Proof for Dilation Invariance (D2)}

\textbf{Statement:} For any positive scalar \(\beta > 0\):
\begin{equation}
    S_{1,2}^*(\beta \Delta W_i) = S_{1,2}^*(\Delta W_i).
\end{equation}

\begin{proof}    
    Consider the matrix after scaling, $\alpha \Delta W_i$. The $\ell_1$ and $\ell_2$ norms are given by:
    
    $$
    \ell_1(\alpha \Delta W_i) = \sum_{a=1}^{m} \sum_{b=1}^{n} |\alpha \Delta W_i[a,b]| = \alpha \ell_1(\Delta W_i)
    $$
    
    \begin{align*}
    \ell_2(\alpha \Delta W_i) &= \left( \sum_{a=1}^{m} \sum_{b=1}^{n} |\alpha \Delta W_i[a,b]|^2 \right)^{1/2} \\
    &= \alpha \ell_2(\Delta W_i)
    \end{align*}

    Substituting these into the sparsity measure formula:
    \begin{align*}
    S_{1,2}^{*}(\alpha \Delta W_i) &= \frac{\sqrt{mn} \cdot \ell_1(\alpha \Delta W_i)}{\ell_2(\alpha \Delta W_i)} \\
    &= \frac{\sqrt{mn} \cdot \alpha \ell_1(\Delta W_i)}{\alpha \ell_2(\Delta W_i)} \\
    &= \frac{\sqrt{mn} \cdot \ell_1(\Delta W_i)}{\ell_2(\Delta W_i)} = S_{1,2}^{*}(\Delta W_i)
    \end{align*}

    \textbf{Conclusion:} Scaling the matrix does not affect the sparsity measure, which satisfies the dilation invariance property (D2).

\end{proof}

\subsubsection{Proof for Rising Tide Property (D3)}

\textbf{Statement:} Adding a positive scalar \(\gamma\) to all elements decreases sparsity:
\begin{equation}
    S_{1,2}^*(\Delta W_i + \gamma) > S_{1,2}^*(\Delta W_i).
\end{equation}

\begin{proof}
    Let \(\Delta W'_i = \Delta W_i + \gamma\).
    
    \textbf{Compute norms:}
    
    \begin{align}
        \Vert \Delta W'_i \rVert_1 &= \Vert \Delta W_i \rVert_1 + mn \gamma, \\
        \Vert \Delta W'_i \rVert_2^2 &= \Vert \Delta W_i \rVert_2^2 + 2\gamma \Vert \Delta W_i \rVert_1 + mn \gamma^2.
    \end{align}
    
    \textbf{Compute sparsity measure:}
    
    \begin{equation}
        S_{1,2}^*(\Delta W'_i) = \frac{\sqrt{mn} \cdot (\Vert \Delta W_i \rVert_1 + mn \gamma)}{\sqrt{\Vert \Delta W_i \rVert_2^2 + 2\gamma \Vert \Delta W_i \rVert_1 + mn \gamma^2}}.
    \end{equation}
    
    Since the numerator increases linearly and the denominator increases sub-linearly, \(S^*_{1,2}(\Delta W'_i) > S^*_{1,2}(\Delta W_i)\), satisfying the Rising Tide Property.
\end{proof}

\subsubsection{Proof for Cloning Invariance(D4)}

\textbf{Statement:} Replicating the matrix should not change its sparsity measure.

\begin{proof}
    Assume that the matrix $\Delta W_i$ is replicated $k$ times to obtain the matrix $\Delta W_i'$:
    
    $$
    \Delta W_i' = [\Delta W_i \mid \Delta W_i \mid \dots \mid \Delta W_i] \quad (k \text{ times})
    $$
    
    The new matrix has dimensions $m \times (kn)$.
    
    The new $\ell_1$ norm is:
    \begin{align*}
    \ell_1(\Delta W_i') &= \sum_{a=1}^{m} \sum_{b=1}^{kn} |\Delta W_i'[a,b]| \\
    &= k \sum_{a=1}^{m} \sum_{b=1}^{n} |\Delta W_i[a,b]| \\
    &= k \ell_1(\Delta W_i)
    \end{align*}

    The new $\ell_2$ norm is:
    \begin{align*}
    \ell_2(\Delta W_i') &= \left( \sum_{a=1}^{m} \sum_{b=1}^{kn} |\Delta W_i'[a,b]|^2 \right)^{1/2} \\
    &= \left( k \sum_{a=1}^{m} \sum_{b=1}^{n} |\Delta W_i[a,b]|^2 \right)^{1/2} \\
    &= \sqrt{k} \ell_2(\Delta W_i)
    \end{align*}
    
    Substituting into the sparsity measure formula:
    
    \begin{align*}
    S_{1,2}^{*}(\Delta W_i') &= \frac{\sqrt{mkn} \cdot \ell_1(\Delta W_i')}{\ell_2(\Delta W_i')} \\
    &= \frac{\sqrt{mkn} \cdot k \ell_1(\Delta W_i)}{\sqrt{k} \ell_2(\Delta W_i)} \\
    &= \frac{\sqrt{mn} \cdot \ell_1(\Delta W_i)}{\ell_2(\Delta W_i)} \\
    &= S_{1,2}^{*}(\Delta W_i)
    \end{align*}
        
    \textbf{Conclusion:} Cloning the matrix does not affect the sparsity measure, satisfying the cloning invariance property (D4).

\end{proof}

\subsubsection{ Proof for Dominance by Maximum Value (P1)}
\textbf{Statement:} When one element becomes very large, sparsity is dominated by that element.
\begin{proof}
    Assume that one element $\Delta W_i[j,k]$ in the matrix becomes infinitely large, while all other elements remain unchanged. Then:
    
    $$
    \ell_1(\Delta W_i) \approx |\Delta W_i[j,k]| + \sum_{(a,b)\neq(j,k)} |\Delta W_i[a,b]|
    $$

    \begin{align*}
    \ell_2(\Delta W_i) &\approx \sqrt{|\Delta W_i[j,k]|^2 + \sum_{(a,b)\neq(j,k)} |\Delta W_i[a,b]|^2} \\
    &\approx |\Delta W_i[j,k]|
    \end{align*}
    
    Thus, the sparsity measure becomes:
    $$
    S_{1,2}^{*}(\Delta W_i) \approx \frac{\sqrt{mn} \cdot |\Delta W_i[j,k]|}{|\Delta W_i[j,k]|} = \sqrt{mn}
    $$
    
    As $\Delta W_i[j,k] \to \infty$, the sparsity measure approaches its maximum value $\sqrt{mn}$.
    
    \textbf{Conclusion:} The sparsity measure is dominated by the largest element, satisfying the Dominance by Maximum Value property (P1).
    
\end{proof}

\subsubsection{Proof for Effect of Adding Zeroes (P2)}
\textbf{Statement:} Adding zeroes increases sparsity.
\begin{proof}
    Assume we add an all-zero column to the matrix $\Delta W_i$, resulting in a new matrix $\Delta W_i'$ with dimensions $m \times (n + 1)$.
    
    The new $\ell_1$ norm is:
    
    \[
    \ell_1(\Delta W_i') = \ell_1(\Delta W_i) + \sum_{a=1}^{m} |0| = \ell_1(\Delta W_i)
    \]
    
    The new $\ell_2$ norm is:
    
    \begin{align*}
    \ell_2(\Delta W_i') &= \left( \sum_{a=1}^{m} \sum_{b=1}^{n+1} |\Delta W_i'[a,b]|^2 \right)^{1/2} \\
    &= \left( l_2(\Delta W_i)^2 + \sum_{a=1}^{m} |0|^2 \right)^{1/2} \\
    &= \ell_2(\Delta W_i)
    \end{align*}

    However, since the dimensions of the matrix have changed from $m \times n$ to $m \times (n + 1)$, the sparsity measure formula becomes:
    
    \begin{align*}
    S_{1,2}^{*}(\Delta W_i') &= \frac{\sqrt{m(n+1)} \cdot \ell_1(\Delta W_i')}{\ell_2(\Delta W_i')} \\
    &= \frac{\sqrt{m(n+1)} \cdot \ell_1(\Delta W_i)}{\ell_2(\Delta W_i)} \\
    &> \frac{\sqrt{mn} \cdot \ell_1(\Delta W_i)}{\ell_2(\Delta W_i)} \\
    &= S_{1,2}^{*}(\Delta W_i)
    \end{align*}

    Since $\sqrt{m(n+1)} > \sqrt{mn}$, we conclude:
    
    \[
    S_{1,2}^{*}(\Delta W_i') > S_{1,2}^{*}(\Delta W_i)
    \]
    
    \textbf{Conclusion:} Adding zero elements increases the sparsity measure, satisfying the Effect of Adding Zeroes property (P2).

\end{proof}

\end{document}